\documentclass[letterpaper, 10 pt, conference, twoside]{ieeeconf}  

\IEEEoverridecommandlockouts                              

\usepackage{amsmath}
\usepackage{subcaption}
\usepackage{multirow}
\usepackage{balance}
\usepackage{hyperref}
\usepackage{dsfont}
\usepackage{breqn}
\usepackage{makecell}
\usepackage{graphicx}
\usepackage{amsfonts}
\usepackage{makecell}

\DeclareCaptionFont{caption}{\fontsize{8}{9.6}\selectfont}
\title{\LARGE \bf From Bench to Bedside: The First Live Robotic Surgery on the dVRK to Enable Remote Telesurgery with Motion Scaling}
\author{Florian Richter$^1$ \IEEEmembership{Student Member, IEEE}, Emily K. Funk$^2$, Won Seo Park$^3$,\\ Ryan K. Orosco$^2$ \IEEEmembership{Member, IEEE}, and Michael C. Yip$^1$ \IEEEmembership{Member, IEEE} 
\thanks{$^1$Florian Richter and Michael C. Yip are with the Department of Electrical and Computer Engineering, University of California San Diego, La Jolla, CA 92093 USA. {\tt\small \{frichter, yip\}@ucsd.edu}}%
\thanks{$^2$Emily K. Funk and Ryan K. Orosco are with the Department of Surgery - Division of Head and Neck Surgery, University of California San Diego, La Jolla, CA 92093 USA. 
{\tt\small \{ekfunk, rorosco\}@ucsd.edu }}%
\thanks{$^3$ Won Seo Park is with the Department of Surgery, School of Medicine, Kyung Hee University, Seoul 02447, Republic of Korea {\tt\small  pwsmd@hanmail.net}}
\vspace{-0.5cm}
}

\begin{document}

\maketitle
\thispagestyle{empty}
\pagestyle{empty}

\begin{abstract}
Innovations from surgical robotic research rarely translates to live surgery due to the significant difference between the lab and a live environment.
Live environments require considerations that are often overlooked during early stages of research such as surgical staff, surgical procedure, and the challenges of working with live tissue.
One such example is the da Vinci Research Kit (dVRK) which is used by over 40 robotics research groups and represents an open-sourced version of the da Vinci\textregistered{} Surgical System.
Despite dVRK being available for nearly a decade and the ideal candidate for translating research to practice on over 5,000 da Vinci\textregistered{} Systems  used in hospitals around the world, not one live surgery has been conducted with it.
In this paper, we address the challenges, considerations, and solutions for translating surgical robotic research from bench-to-bedside.
This is explained from the perspective of a remote telesurgery scenario where motion scaling solutions previously experimented in a lab setting are translated to a live pig surgery.
This study presents results from the first ever use of a dVRK in a live animal and discusses how the surgical robotics community can approach translating their research to practice.
\end{abstract}

\section{Introduction}

Since the advent of surgical robotics, a paradigm shift occurred in modern surgery, and many procedures have been converted from traditional open approaches to endoscopic or minimally invasive approaches  \cite{mack2001minimally}.
The use of robotic technologies in the operating room allows the surgeon to provide target treatment through a minimally invasive approach, but also removes the need for the surgeon to be physically next to the patient undergoing the procedure.
Instead, the surgeon operates the robotic device from a console.
Telesurgery systems allow for advanced forms of manipulation in confined spaces with unique robotic designs and potentially higher precision through robotic instrumentation.
For example, Intuitive's da Vinci\textregistered{} Surgical System providers more degrees of freedom than standard hand-held laparoscopic tools and reduces physiologic tremor \cite{kwartowitz2006toward}.

The introduction of robotic technology to the operating environment has fostered a substantial expansion in surgical robotic research from engineering groups, including areas of surgical perception \cite{li2020super} and task automation \cite{yipDasJournal}.
Despite the immense work that is currently being conducted in this field, there is a paucity of innovative designs that have successfully been translated and applied in the clinical context.
The coordination and collaboration that is required to translate innovative designs from the laboratory setting into applications that can be evaluated in live surgical experiments represents a major obstacle.
Too often this barrier prevents the innovative designs of an engineering team from making into clinical trials or application.
Additionally, introducing a novel engineered design into a live environment requires clinical consideration that are often overlooked in the early stages of research such as surgical procedure, surgical staff, equipment and facilities that allow for live surgery research, and consideration of the challenges and differences that occur when working with live tissue.
To this end, few research groups have ever managed to reach live animal or human trials \cite{rentschler2004vivo,wortman2011laparoendoscopic, csen2016system}.

\begin{figure}[t]
    \centering
    \vspace{1mm}
    \includegraphics[ width=0.23\textwidth]{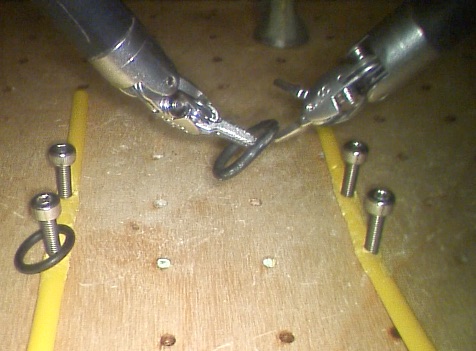}
    \includegraphics[ width=0.23\textwidth]{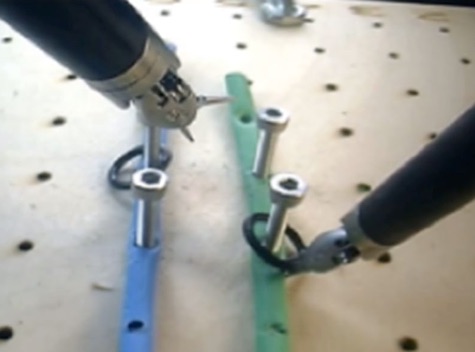}
    
    \vspace{1mm}
    
    \includegraphics[ width=0.4675\textwidth]{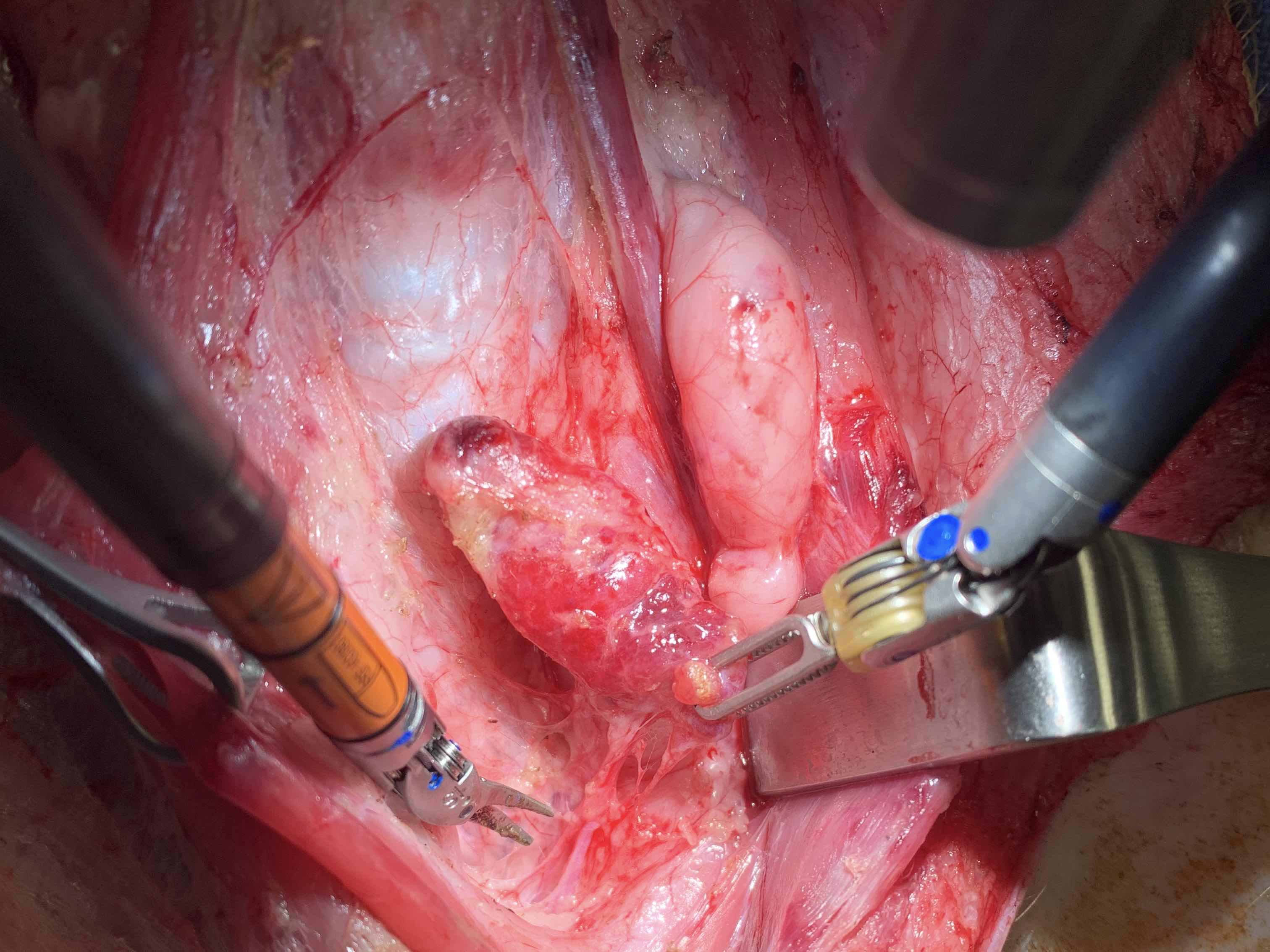}
    \caption{From bench to bedside: the top two figures are from previous laboratory experiments \cite{richter2019motion, orosco2020compensatory} and bottom figure is from an in-vivo thyroidectomy on a porcine model with dVRK. The experiments are conducted under delay to simulate a remote telesurgery scenario, and motion scaling solutions are applied to overcome the deleterious effects of signal latency.}
    \label{fig:cover_figure}
\end{figure}

An example of a research platform that has yet to show translation to live surgery is the research kit version of the Intuitive's da Vinci\textregistered{} Surgical System, the da Vinci Research Kit (dVRK) \cite{kazanzides2014open}.
This research platform is used by over 40 institutions around the world and yielded over 250 publications and preprints since its conception 10 years ago \cite{d2021accelerating}.
With over 5,000 da Vinci\textregistered{} Systems being used in hospitals globally \cite{intuitive}, the dVRK is an ideal candidate for effectively deploying surgical robotic research efforts to live surgery.
However, despite nearly a decade of availability and hundreds of published papers, not one live surgery has been conducted with a dVRK.

In this paper, we address the challenges, considerations, and solutions for translating surgical robotics research from bench-to-bedside.
We walk through the use of a dVRK in a remote telesurgery setting, describing how innovations in control, previously established in the lab, can be translated to and tested in a live animal setting.
This study presents results from the first ever use of a dVRK in a live animal and discusses how the dVRK and greater surgical robotics community can approach translating research towards clinical practice.
We will present our established solution in the field of telesurgery and describe the required efforts that allowed our translation of this solution from lab experiments to live surgery.
This work will serve as a model to establish an outline of the rigorous steps taken to ultimately achieve in-vivo results.
In addition, we will highlight pertinent specific challenges during this process to help further expand the understanding of all the required considerations in the preparation of this type of work.

\section{Motion Scaling for Remote Telesurgery}

Remote surgery, or telesurgery, describes the ability of a surgeon to utilize a robotic platform to operate on a patient in a different physical location.
This alluring idea would allow people in remote parts of the country or even military personnel across the globe to receive surgical treatment from highly experienced surgeons.
This concept was introduced along with the robotic platforms over a decade ago and was thought to be well within reach given the teleoperational nature of the robotic systems, which is displacing the surgeon physically from the patient's bedside.
The actual implementation and use of telesurgery, however, has not been as successful.
A teleoperation system has telecommands sent from the \textit{remote-administrator} to the \textit{bed-side-manipulator}.
Then feedback from the bed-side-manipulator, such as endoscopic camera data, is sent back to the operator at the remote-administrator.
The feedback gives the perception necessary for the surgeon to close the loop and achieve complex forms of manipulation.
However, latency in the communication channel which transports the telecommands and feedback causes deleterious effects such as oscillations \cite{SteadyHand, TreatmentPlanning}, hence impeding the surgeon's ability.
The unavoidable signal latency only grows as the distance between the surgeon and bedside increases, and previously it has been measured to be over 500msec when using satellite communication between London and Toronto \cite{reasonForDelay}.

\begin{figure}
    \centering
    \vspace{2mm}
    \includegraphics[trim=1cm 2.15cm 1cm 1.35cm, clip, width=\linewidth]{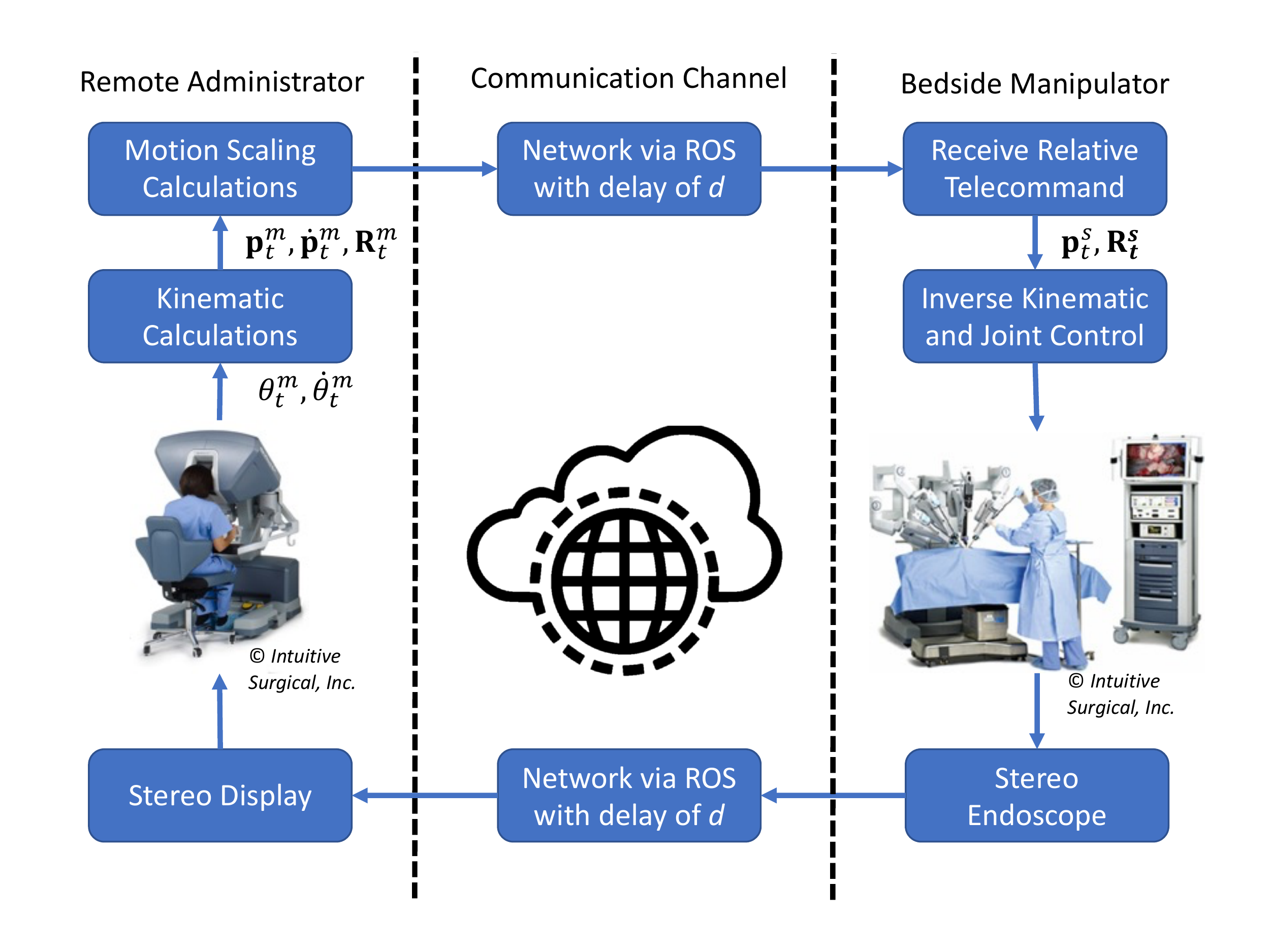}
    \caption{A block diagram of implementing motion scaling solutions on dVRK for remote telesurgery. The telecommands and feedback are sent over network via ROS which causes a delay. To overcome the deleterious effects of the delay, motion scaling is applied to the telecommands before sending to the bedside manipulator.}
    \label{fig:flow_chart}
\end{figure}

Previously, we investigated motion scaling techniques to help alleviate the challenges in performing complex manipulations when teleoperating surgical robots under delay \cite{richter2019motion, orosco2020compensatory}.
A flowchart of the motion scaling solutions are shown in Fig. \ref{fig:flow_chart}.
Its previous lab experimentation results shows promise to being directly effective in live surgeries because of its simplicity in implementation and it applies no additional limitations to the manipulations that can be performed by the surgeon.
Therefore, the next big step in experimentation towards deployment and enabling remote telesurgery is conducted for motion scaling, live surgery.

Motion scaling solutions modify the telecommand sent form the remote-administrator to the bed-side-manipulator.
The remote-administrator in surgical robots is typically a dexterous back-drivable manipulator such as the Master Tool Manipulators (MTM) on a da Vinci\textregistered{} Surgical Robot.
The dexterity allows the operator to input complex trajectories in an intuitive manner which the bed-side-manipulator will follow.
Let the remote-administrators position and rotation matrix at time $t$ be denoted as $\mathbf{p}^m_t \in \mathbb{R}^3$ and $\mathbf{R}^m_t \in SO(3)$ respectively in its own base frame.
Then the target position of the remote manipulator with a one way delay of $d$ is:
\begin{align}
    \label{eq:position_set_point}
    \mathbf{p}^s_t &=  \mathbf{R}^{s}_{m} \gamma ( \mathbf{p}^m_{t-d} -  \mathbf{p}^m_{t-d-1}) + \mathbf{p}^s_{t-1} \\
    \label{eq:orientation_set_point}
    \mathbf{R}^s_t &= \mathbf{R}^{s}_{m} \mathbf{R}^m_{t-d}
\end{align}
where $\mathbf{p}^s_t \in \mathbb{R}^3$ and $\mathbf{R}^s_t \in SO(3)$ are the bed-side-manipulator's position and rotation matrix respectively in its own base frame at time $t$, $\mathbf{R}^{s}_{m} \in SO(3)$ is a rotation matrix that aligns the remote-administrator with the bed-side-manipulator, and $\gamma$ is a scaling factor.

The position command shown in (\ref{eq:position_set_point}) is designed to send relative telecommands to bed-side-manipulator.
This way the remote-administrator's positional workspace does not need to be one-to-one with the bed-side-manipulator's.
It also allows for clutching which is when the operator moves the remote-administrator manipulator without sending commands, hence allowing the operator to reset their local positional workspace.
Meanwhile the orientation command shown in (\ref{eq:orientation_set_point}) is kept one-to-one due to the lack of intuitiveness in operator control when there is misalignment in orientation.

The scaling factor from (\ref{eq:position_set_point}), $\gamma$, is modified for the motion scaling solutions.
Overall, it is decreased to slow down the bed-side-manipulator's motions.
This is similar to the move and wait strategy where operators create stability in a teleoperated system under delay by moving only after observing a response in the feedback \cite{MoveAndWait}.
However, here the motions are slowed continuously, and the move and wait strategy requires operators to have experience teleoperating under delay.
Of course, decreasing the scaling factor will increase the time to complete tasks.
Therefore, the scaling factor is dynamically adjusted as follows:
\begin{equation}
    \gamma = \gamma_c + \gamma_v || \dot{\mathbf{p}}^m_{t-d} ||
    \label{eq:velocity_scaling}
\end{equation}
where $\gamma_c$ and $\gamma_v$ are the constant and velocity scaling factors respectively and $|| \cdot ||$ is the $l_2$ norm.
The velocity scaling term is similar to mouse acceleration on a computer and is based on the idea that the operator will naturally move slower when higher precision is required.
So (\ref{eq:velocity_scaling}) has the low base-scaling, set by $\gamma_c$, during precise motions and will dynamically increase when lower precision is needed.

To calculate the telecommands, let $\theta^m_t \in \mathbb{R}^n$ be the remote-administrators $n$ joint angle readings at time $t$.
Then its pose, $\mathbf{p}^m_t, \mathbf{R}^m_t$, is calculated using forward kinematics, $f(\theta^m_t)$, and its linear velocity is calculated by:
\begin{equation}
    \mathbf{\dot{p}}^m_{t} = \mathbf{J} \dot{\theta}^m_t
\end{equation}
where $\mathbf{J} \in \mathbb{R}^{3 \times n}$ is the Jacobian of $f(\cdot)$ with respect to the position, $\mathbf{p}^m_t$.
The calculations for the telecommand are implemented directly in dVRK where $f(\cdot)$ and $\mathbf{J}$ are defined for the MTMs and $\theta^m_t$ and $\dot{\theta}^m_t$ are measured by dVRK's motor controller systems.
The relative telecommand is then sent over network via the Robot Operating System (ROS) using dVRK's ROS bridge \cite{chen2017software} to the bedside-manipulator.
The bedside-manipulator is dVRK's Patient Side Manipulator (PSM) which is regulated to the target pose command, $\mathbf{p}^s_t$ and $\mathbf{R}^s_t$.

The feedback for the surgeon is stereo-scopic data from an Endoscopic Camera Manipulator (ECM).
Similar to the telecommands, it is sent over network via ROS to an image viewer.
The image viewer displays the left and right video data to the dVRK's master console which supplies stereo viewing to the operator.
By using network as the communication channel, the system can be separated to conduct remote teleoperation.
For the experimentation in this work, an artificial delay is added in the ROS communication channel to both the telecommands and visual feedback in order to simulate remote telesurgery.

\begin{figure}
    \centering
    \vspace{2mm}
    \includegraphics[ width=\linewidth]{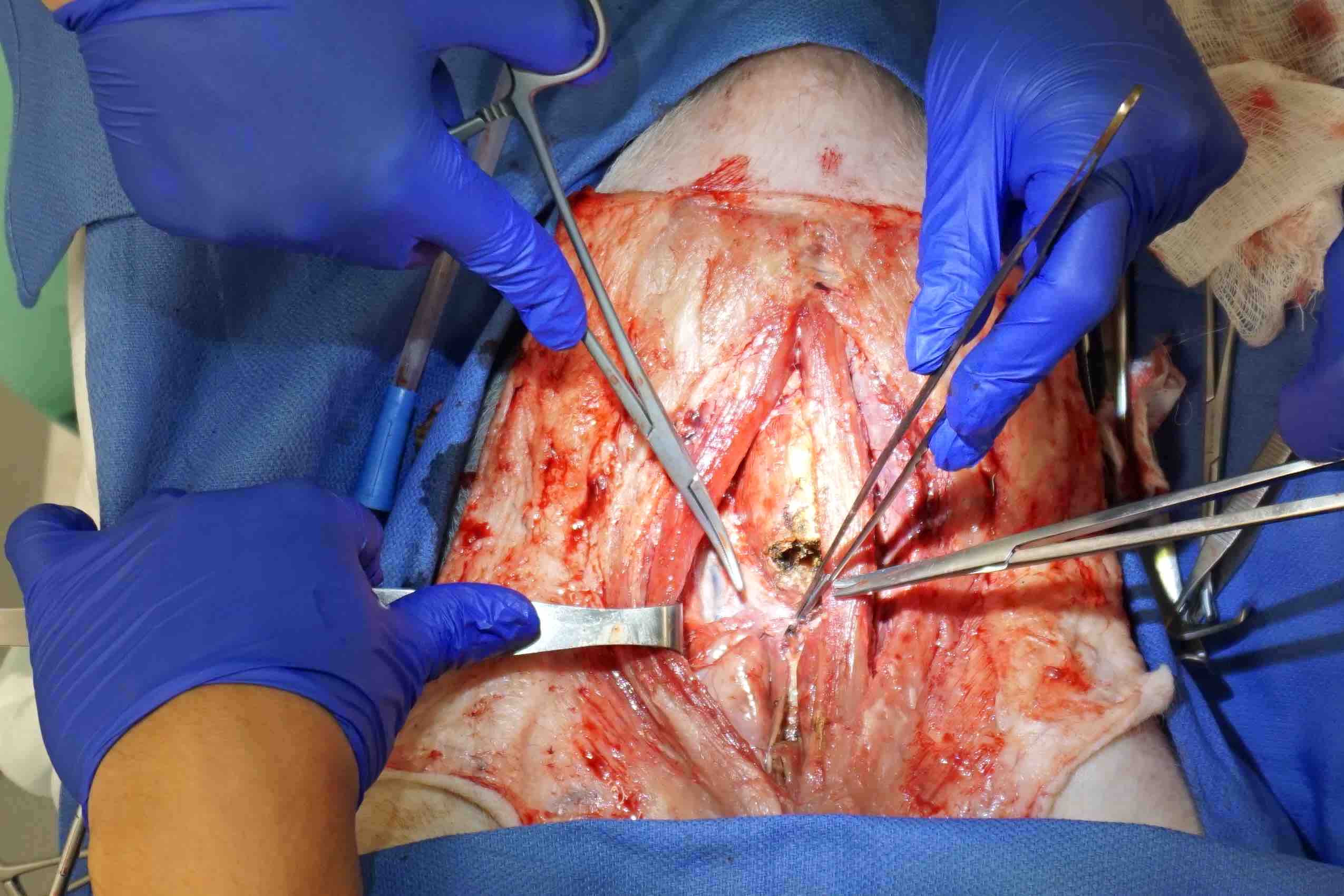}
    \caption{Manual dissection of pig cadaver to familiarize the surgical team with porcine anatomy. This understanding was used to craft the surgical steps taken for the in-vivo motion scaling experiment. }
    \label{fig:manual_dissection}
\end{figure}

\begin{figure*}
    \centering
    \vspace{2mm}
    \includegraphics[width=0.24\linewidth]{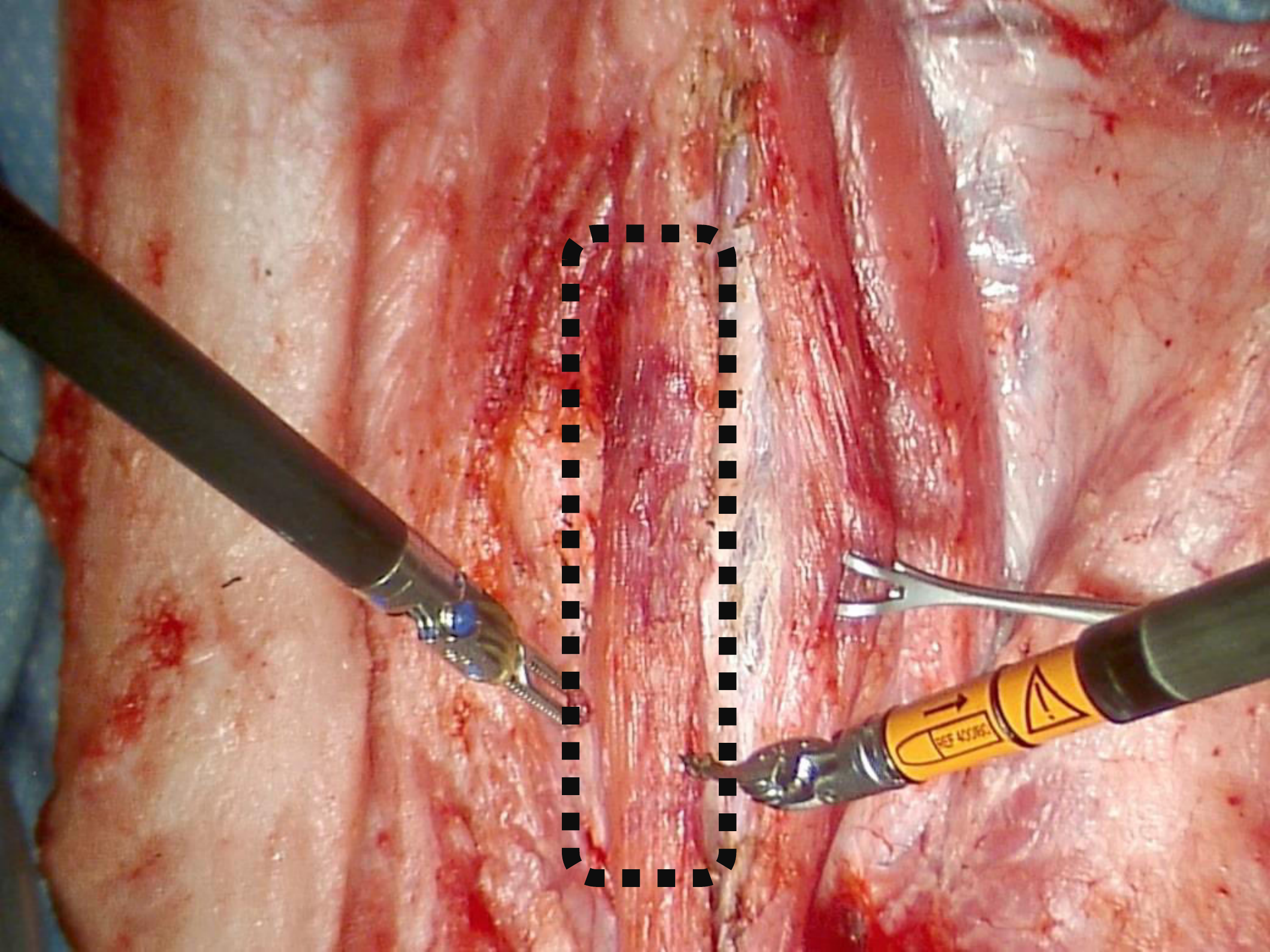}
    \includegraphics[width=0.24\linewidth]{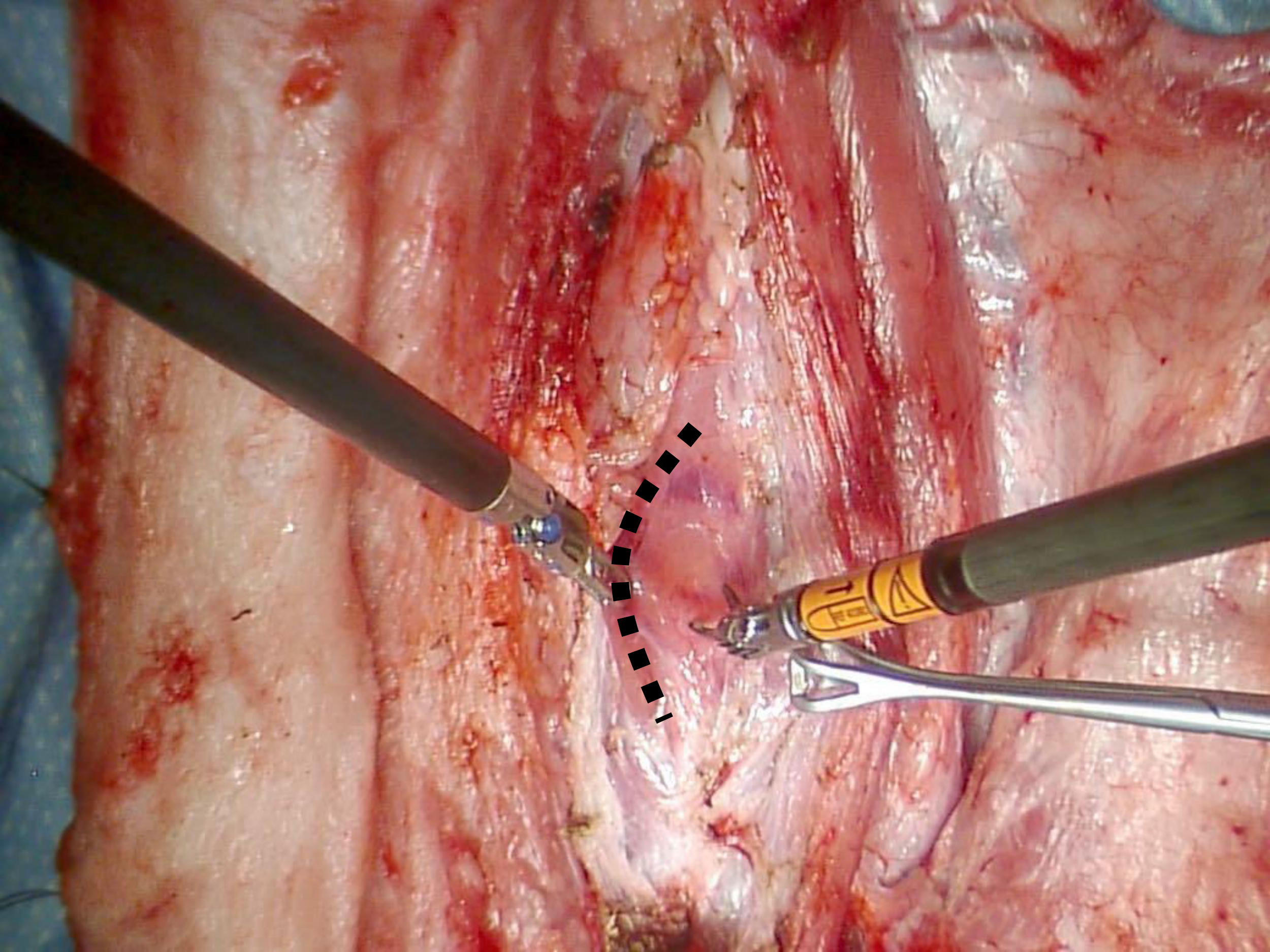}
    \includegraphics[ width=0.24\linewidth]{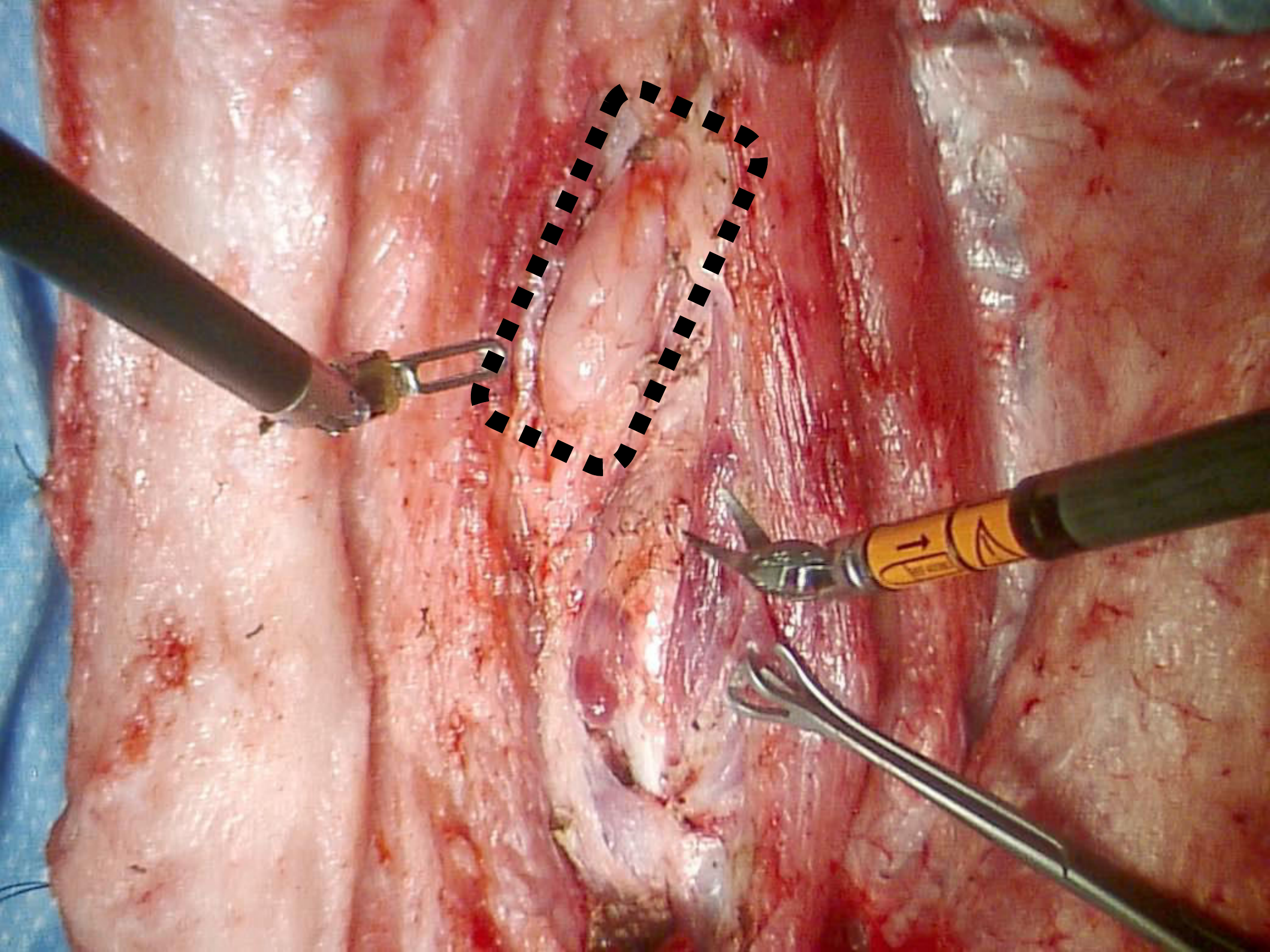}
    \includegraphics[ width=0.24\linewidth]{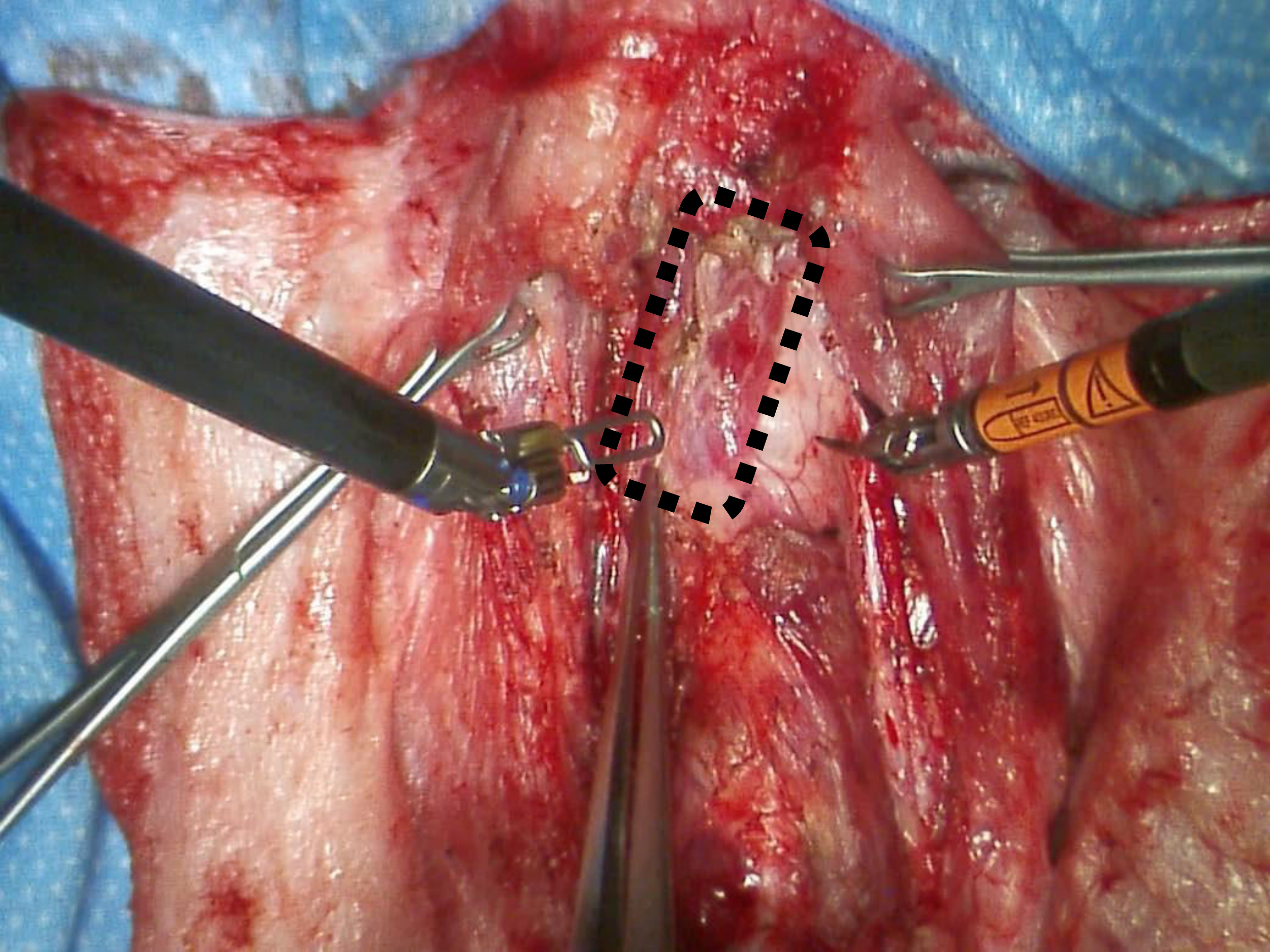}    
    \vspace{1mm}
    
    \includegraphics[width=0.24\linewidth]{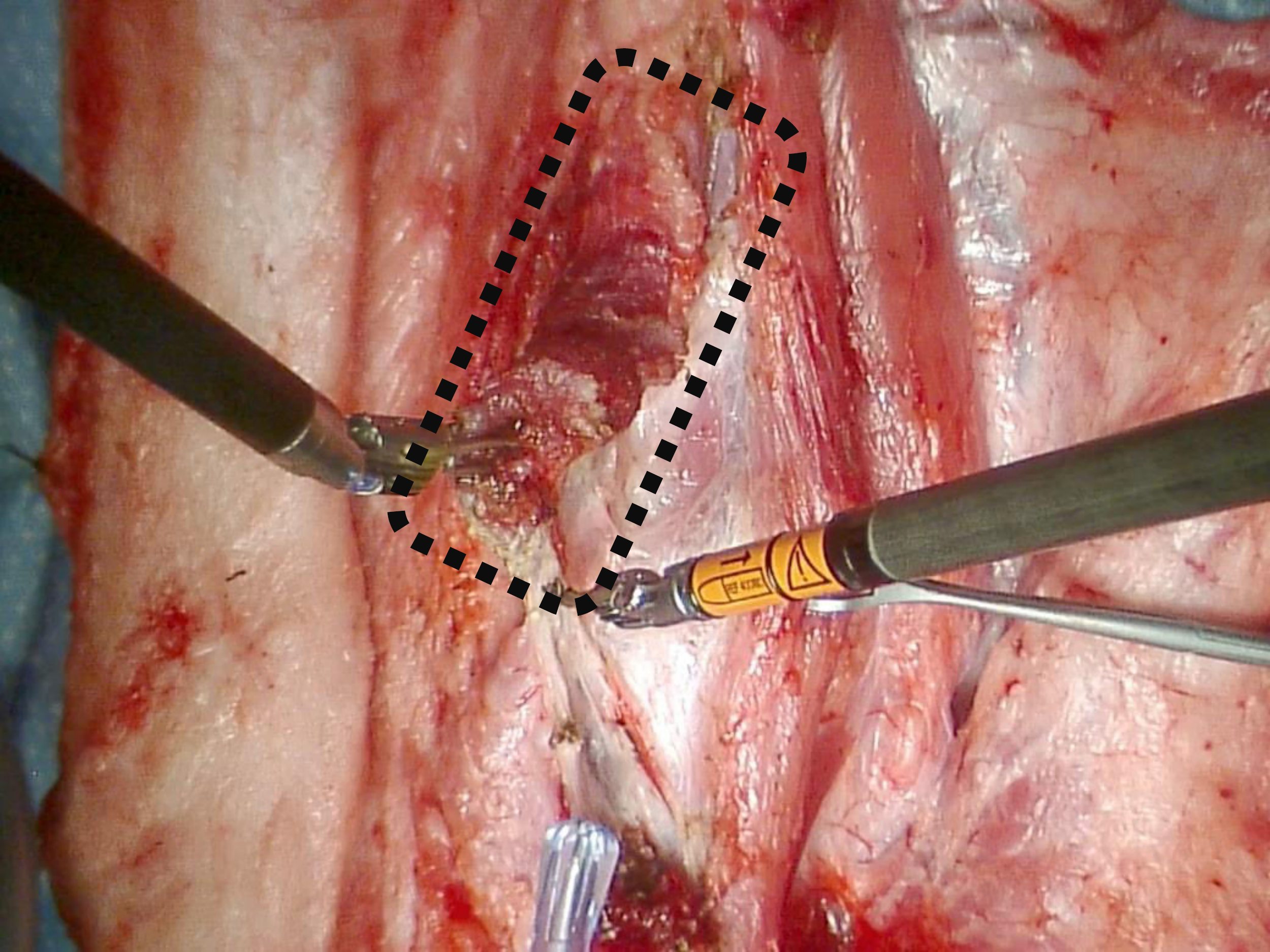}
    \includegraphics[width=0.24\linewidth]{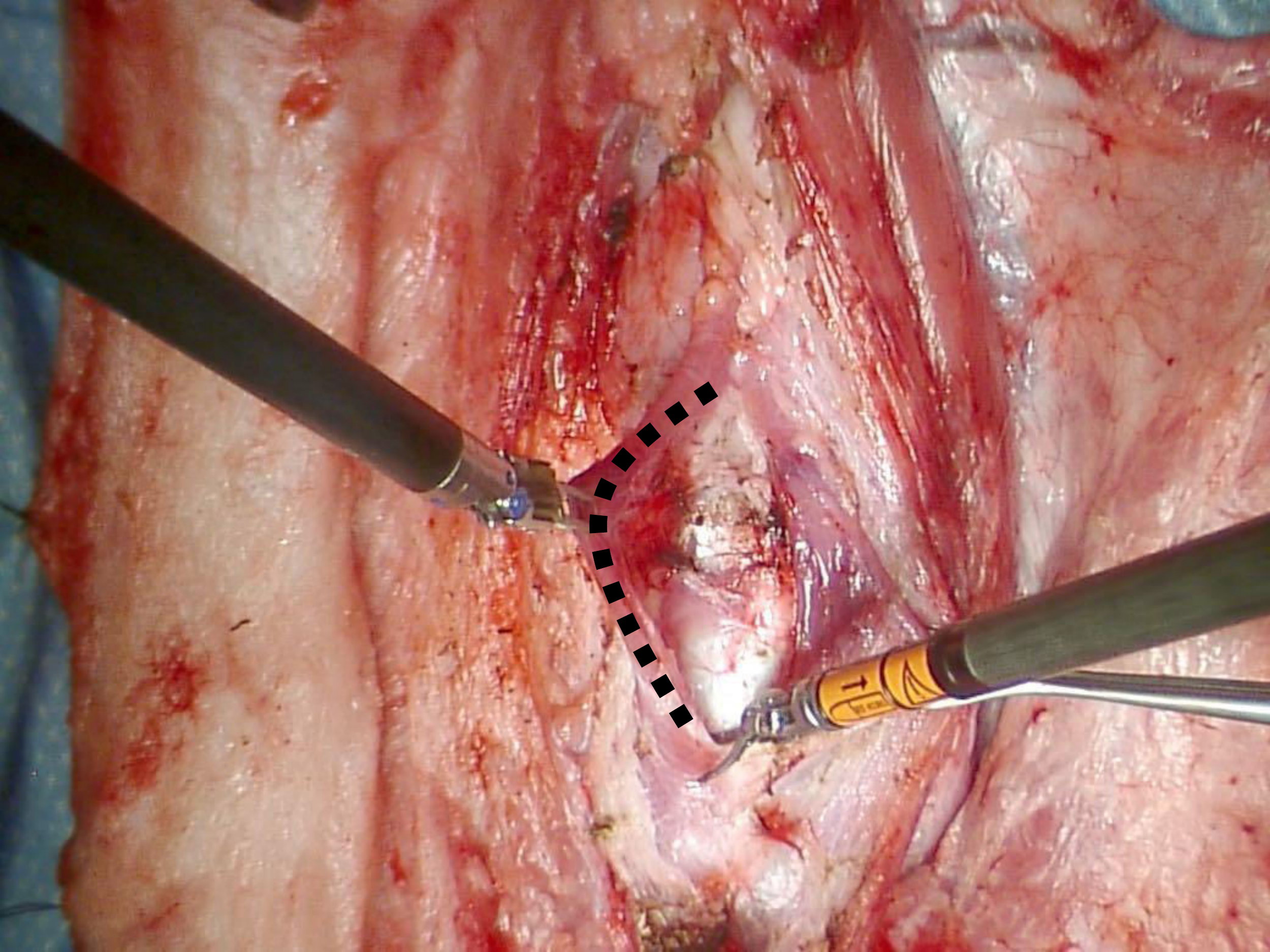}
    \includegraphics[ width=0.24\linewidth]{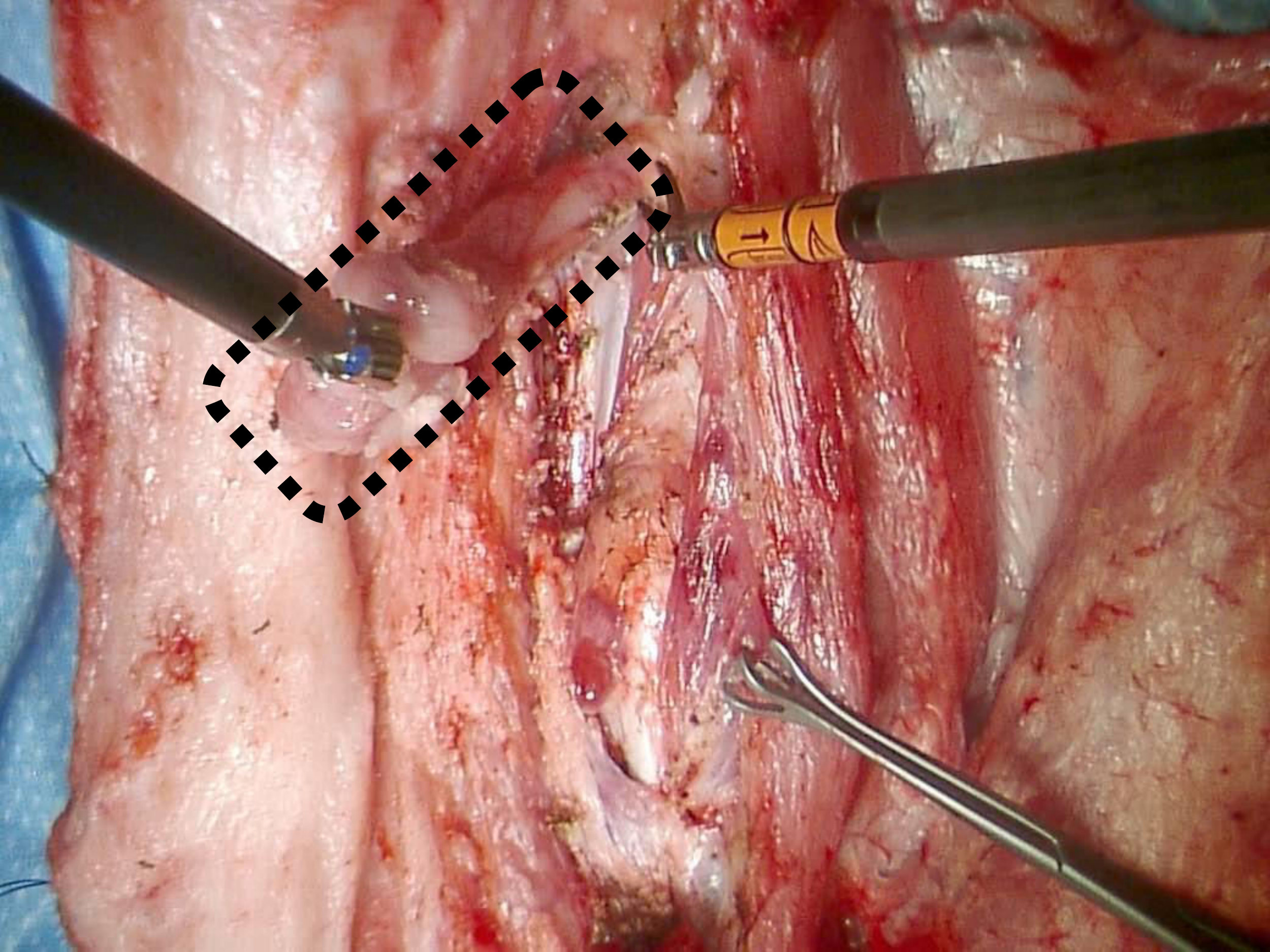}
    \includegraphics[ width=0.24\linewidth]{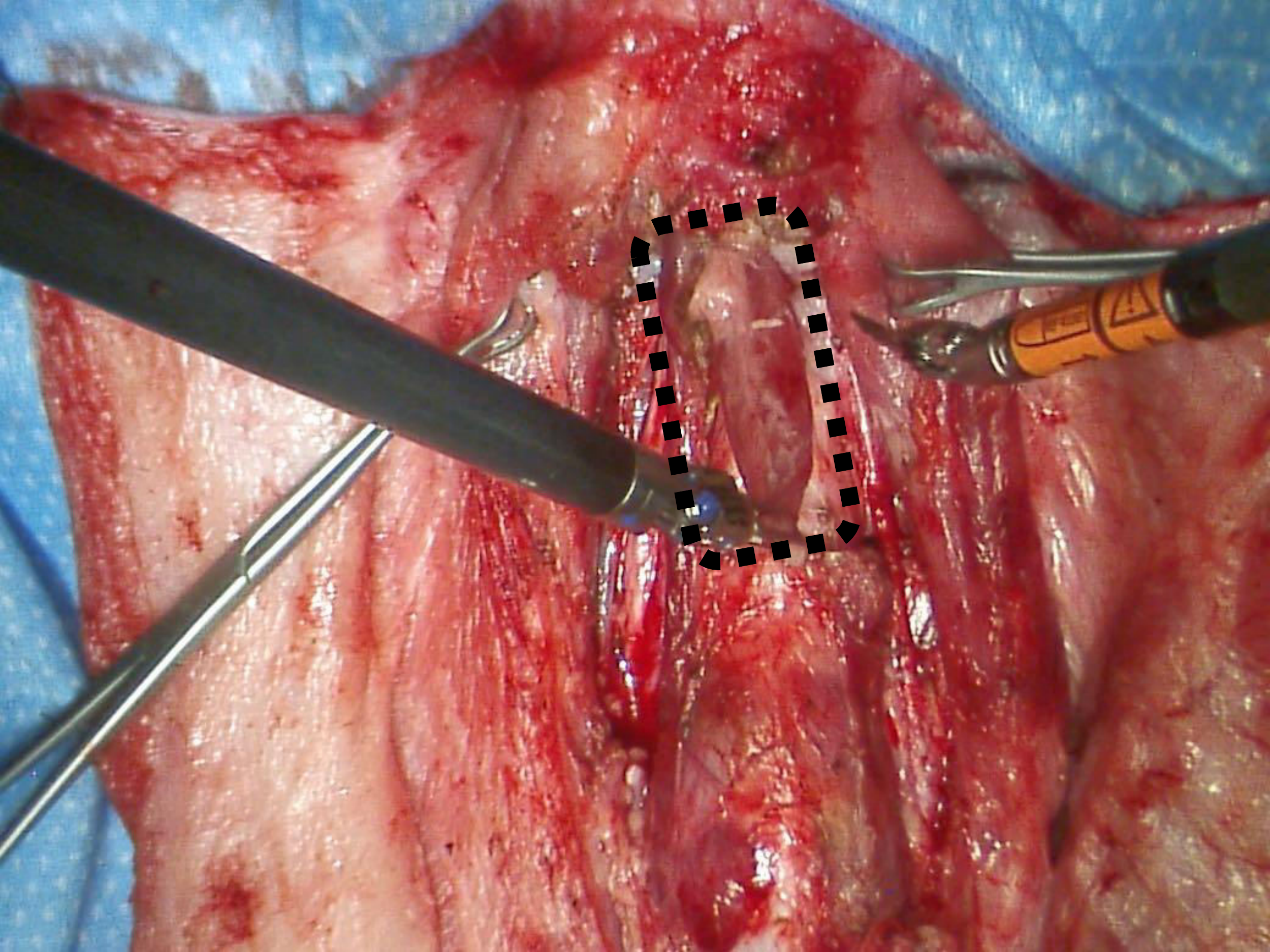}
    \caption{These are the four steps completed in the live surgery to evaluate motion scaling on dVRK. From left to right, the columns show resections of the following anatomy: sternohyoid, sternothyroid, thymus, and thyroid. The highlighted region in each figure is the corresponding anatomy being resected.}
    \label{fig:surgical_steps}
\end{figure*}

\section{Procedure Selection and Animal Model}

Development of an in-vivo trial requires the selection of a specific surgical procedure to be performed and an appropriate animal model for the procedure.
Given that the surgical team is comprised of otolaryngologists (i.e. Head and Neck surgeons), a thyroidectomy is elected as the surgery for evaluation of the motion scaling solutions.
Thyroidectomy involves the excision of the thyroid gland in the anterior neck for treatment of benign and malignant disorders.
This procedure can be performed in various manners including open non-robotic approaches or through minimally invasive techniques with a robotic platform \cite{chae2020comparison}.
Furthermore, the surgical steps and anatomy have little variability between cases making it easier to compare between multiple trials.
After review of previous literature \cite{wu2010investigation, zhang2019feasibility}, the porcine model was selected as it is the most used animal model for this procedure.
Additionally, the pig model is economical, widely available, and offers anterior neck anatomy similar to that of humans.

Initially, two open dissections without robotic assistance were performed to familiarize the surgical team with the neck anatomy of the pig.
These dissections highlighted pertinent differences between the model and human anatomy and helped guide further surgical planning.
With this experience, an adapted sequence of steps were developed for porcine thyroidectomy that would maximize the utilization of each animal and procedure.
The purpose of doing this was to collect more data per pig for evaluation of motion scaling.
Therefore, the final procedure resulted in additional surgical steps which are not done in a standard thyroidectomy.

The full list of surgical steps to be completed on a single pig using the dVRK is resection of the following anatomy:
\begin{enumerate}
    \item Left and right sternohyoid
    \item Left and right sternothyroid
    \item Left and right thymus
    \item Thyroid
\end{enumerate}
The described anatomy and steps are shown in Fig. \ref{fig:surgical_steps}.
To efficiently split the steps for the telesurgery study, the procedure per pig is split into two parts. Part one is resections of left sternohyoid, sternothyroid, and thymus for one set of data, and part two is resection of right sternohyoid, sternothyroid, thymus and thyroid for a second set of data.

\section{Final Adjustments Before Live Surgery}

An additional two cadaveric surgeries were performed to test the procedure with the dVRK and troubleshoot any bugs or failures.
This also allows the surgeons to familiarize themselves with the dVRK system, which has pertinent dissimilarities to the Intuitive platforms currently used in hospitals.
The proposed surgical procedure sequence was also evaluated to ensure all substeps would be consistent between each surgery for comparison.
In addition, these procedures helped determine what additional supplies and equipment were required prior to live procedures.
This included surgical instruments, suction for liquid and smoke evacuation, electrocautery and other miscellaneous supplies.
The surgical team also decided to use the the Fenestrated Forceps in the left-hand bedside-manipulator for grasping and the Monopolar Maryland Bipolar Scissors in the right-hand bedside-manipulator for cautery and cutting.

The two cadaveric surgeries also highlighted two potentially inconsistent variables in the procedures, the bedside assistant and ECM movement.
Overall, the goal is to study the effects motion scaling has on controlling the bedside-manipulators, so minimizing any additional intervention in the trails is necessary for consistent results.
Therefore, the bedside assistant was limited to only helping with:
\begin{itemize}
    \item suction of blood and smoke
    \item applying minimal amounts of surgical clamps for tension
    \item cleaning of instrumentation
\end{itemize}
Meanwhile the ECM remained stationary during the entire procedure.
Although these constraints increase the difficulty of the procedure, it helps isolate the effect motion scaling has on remote telesurgery which is the goal of this experimentation.

\begin{figure}
    \centering
    \includegraphics[trim={0cm 0cm 0cm 16cm}, clip, width=\linewidth]{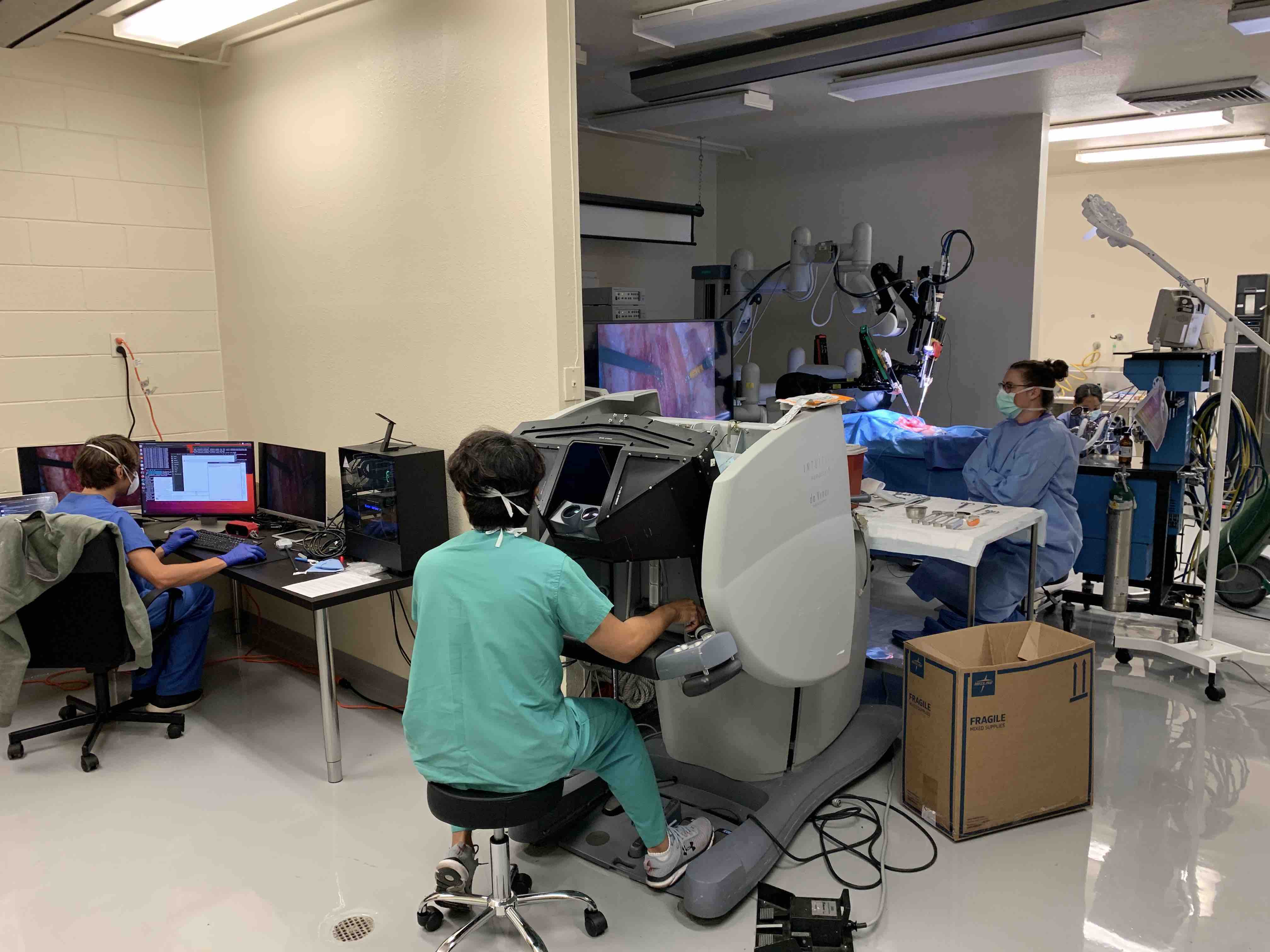}
    \caption{Photo of live surgery in action. The surgeon is at the remote-administrator console to send telecommands to the bed-side-manipulators in the pig. A bedside assistant and anaesthesiologist are next to the pig to monitor the progress of the surgery and assist when necessary. Meanwhile, an engineer monitors the status of the network communication channel, remote-administrator, and bed-side manipulators. }
    \label{fig:operation_room}
\end{figure}

\section{In-Vivo Experimentation and Results}


Two pigs are used to get a full set of experimental results.
The breakdown of various delay and motion scaling scenarios, based on our previous work \cite{richter2019motion, orosco2020compensatory}, are:
\begin{enumerate}
    \item Left resections and no thyroid on pig 1 with no added delay and normal constant scaling: $\gamma_c = 0.30$, $\gamma_v = 0.0$
    \item Right resections and thyroid on pig 1 with added delay and normal constant scaling: $\gamma_c = 0.30$, $\gamma_v = 0.0$
    \item Left resections and no thyroid on pig 2 with added delay and reduced constant scaling: $\gamma_c = 0.15$, $\gamma_v = 0.0$
    \item Right resections and thyroid on pig 2 with added delay and velocity scaling: $\gamma_c = 0.15$, $\gamma_v = 0.1$
\end{enumerate}
Step 1 of this experimental procedure is done to give the surgeon time to adjust to the procedure and dVRK.
Meanwhile step 2, 3, and 4 have the added delay to simulate remote telesurgery and test the effects of motion scaling.
Step 3 and 4 test reducing the constant scaling and applying velocity scaling respectively.
In our experiments, the round-trip delay is set to 500msec, so the one-way delay, $d$, is 250msec.


\begin{table}
\vspace{2mm}
\begin{center}
\def\arraystretch{1.25}
\setlength\tabcolsep{0.45em}
\caption{Time to finish surgical resection of the respective anatomy under a round-trip delay of 500msec with various scaling configurations.}
\begin{tabular}{ c|c c c c}
\textbf{Scaling}
  & \textbf{Sternohyoid} & \textbf{Sternothyroid}  & \textbf{Thymus} & \textbf{Thyroid} \\ \hline 
$ \gamma_c = 0.30, \gamma_v = 0.0 $ &  694sec & 185sec &  903sec & 259sec \\
 $ \gamma_c = 0.15, \gamma_v = 0.0 $ & 1233sec & 759sec & 1078sec & --- \\
 $ \gamma_c = 0.15, \gamma_v = 0.1 $ & \textbf{684sec} &\textbf{166sec} & \textbf{785sec} & \textbf{201sec}
\end{tabular}
\label{table:time_to_finish}
\end{center}

\begin{center}
\def\arraystretch{1.25}
\setlength\tabcolsep{0.45em}
\caption{Total distance travelled by left-hand bedside-manipulator to finish surgical resection of the respective anatomy under a round-trip delay of 500msec with various scaling configurations.}
\begin{tabular}{ c|c c c c}
\textbf{Scaling}
  & \textbf{Sternohyoid} & \textbf{Sternothyroid}  & \textbf{Thymus} & \textbf{Thyroid} \\ \hline 
 $ \gamma_c = 0.30, \gamma_v = 0.0 $  & 3.16m & 0.54m & 3.12m & 0.9m \\
 $ \gamma_c = 0.15, \gamma_v = 0.0 $& 4.02m & 1.60m & 2.5m & --- \\
  $ \gamma_c = 0.15, \gamma_v = 0.1 $ & \textbf{1.72m} & \textbf{0.42m} & \textbf{2.74m} & \textbf{0.44m}
\end{tabular}
\label{table:distance_l_to_finish}
\end{center}

\begin{center}
\def\arraystretch{1.25}
\setlength\tabcolsep{0.45em}
\caption{Total distance travelled by right-hand bedside-manipulator to finish surgical resection of the respective anatomy under a round-trip delay of 500msec with various scaling configurations.}
\begin{tabular}{ c|c c c c}
\textbf{Scaling}
  & \textbf{Sternohyoid} & \textbf{Sternothyroid}  & \textbf{Thymus} & \textbf{Thyroid} \\ \hline 
  $ \gamma_c = 0.30, \gamma_v = 0.0 $  & 4.47m & 1.08m & 6.45m & 1.82m\\
  $ \gamma_c = 0.15, \gamma_v = 0.0 $ & 5.02m & 2.78m & 4.09m & --- \\
  $ \gamma_c = 0.15, \gamma_v = 0.1 $ & \textbf{2.54m} & \textbf{0.57m} & \textbf{3.89m} & \textbf{0.88m}
\end{tabular}
\label{table:distance_r_to_finish}
\end{center}
\end{table}

Kinematic and endoscopic camera data is recorded for all experiments.
The total time and distance travelled by bedside-manipulators of the surgical steps are shown in Tables \ref{table:time_to_finish} - \ref{table:distance_r_to_finish}.
On average, reducing the constant scaling and using velocity scaling modified the time to complete the surgical tasks under delay by 175\% and 88\% respectively relative to normal constant scaling.
Meanwhile, reducing the constant scaling and using velocity scaling on average modified the total distance travelled by the bed-side manipulators by 156\% and 61\% respectively relative to normal constant scaling.
Furthermore, velocity scaling resulted in the lowest times and shortest distance travelled by bedside-manipulators for all surgical tasks.

\begin{figure*}
    \centering
    \vspace{2mm}
    \includegraphics[width=0.32\linewidth]{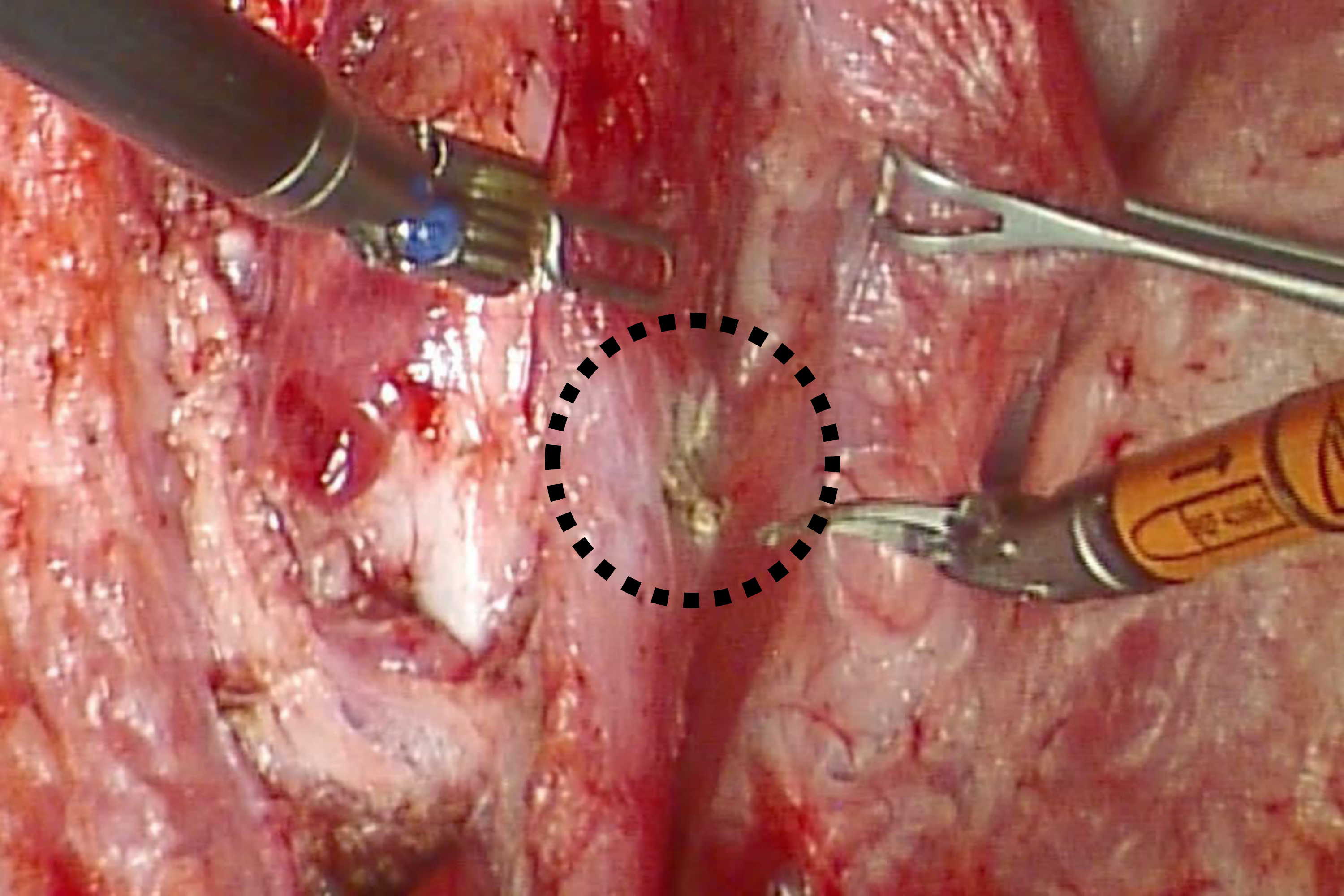}
    \includegraphics[width=0.32\linewidth]{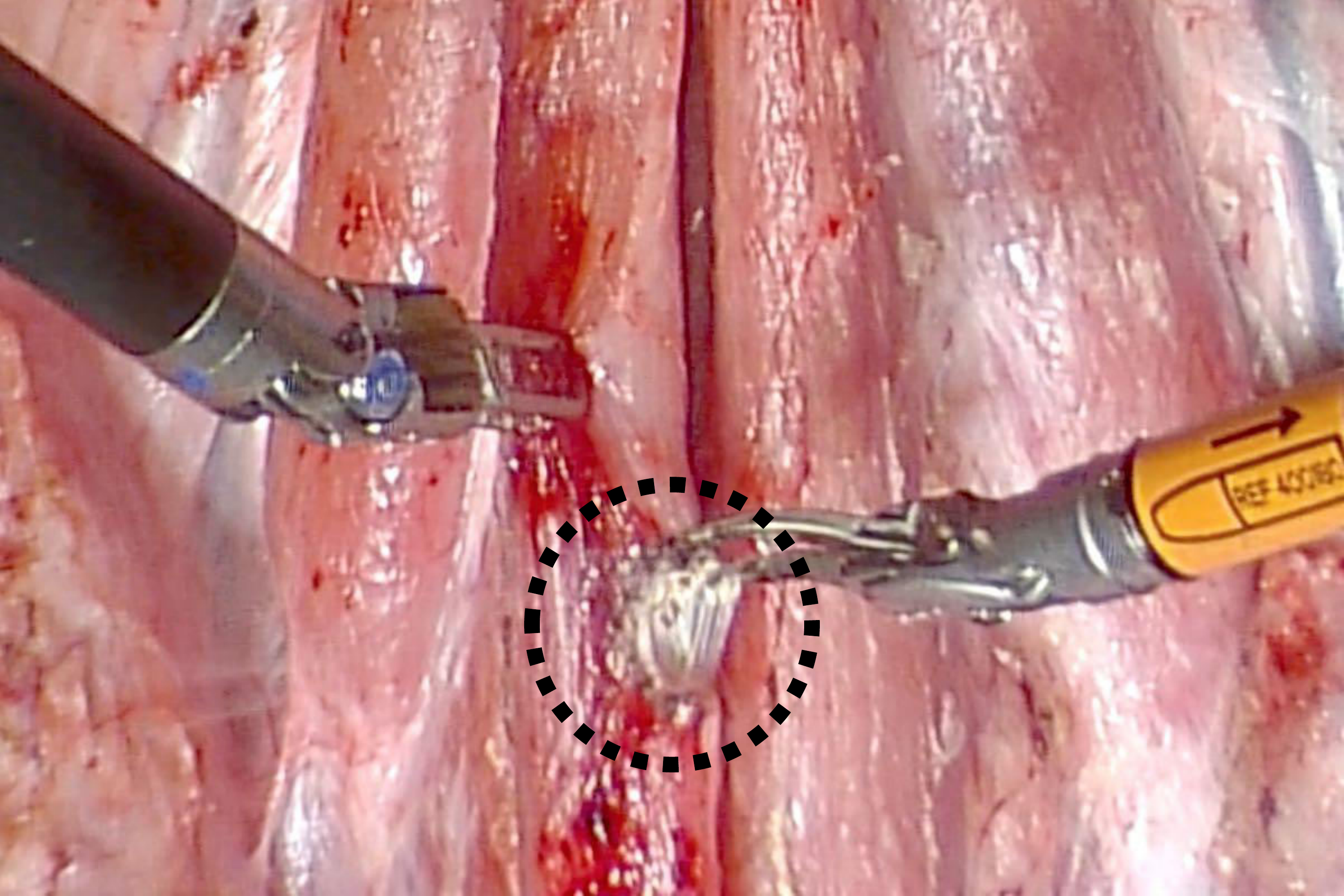}
    \includegraphics[width=0.32\linewidth]{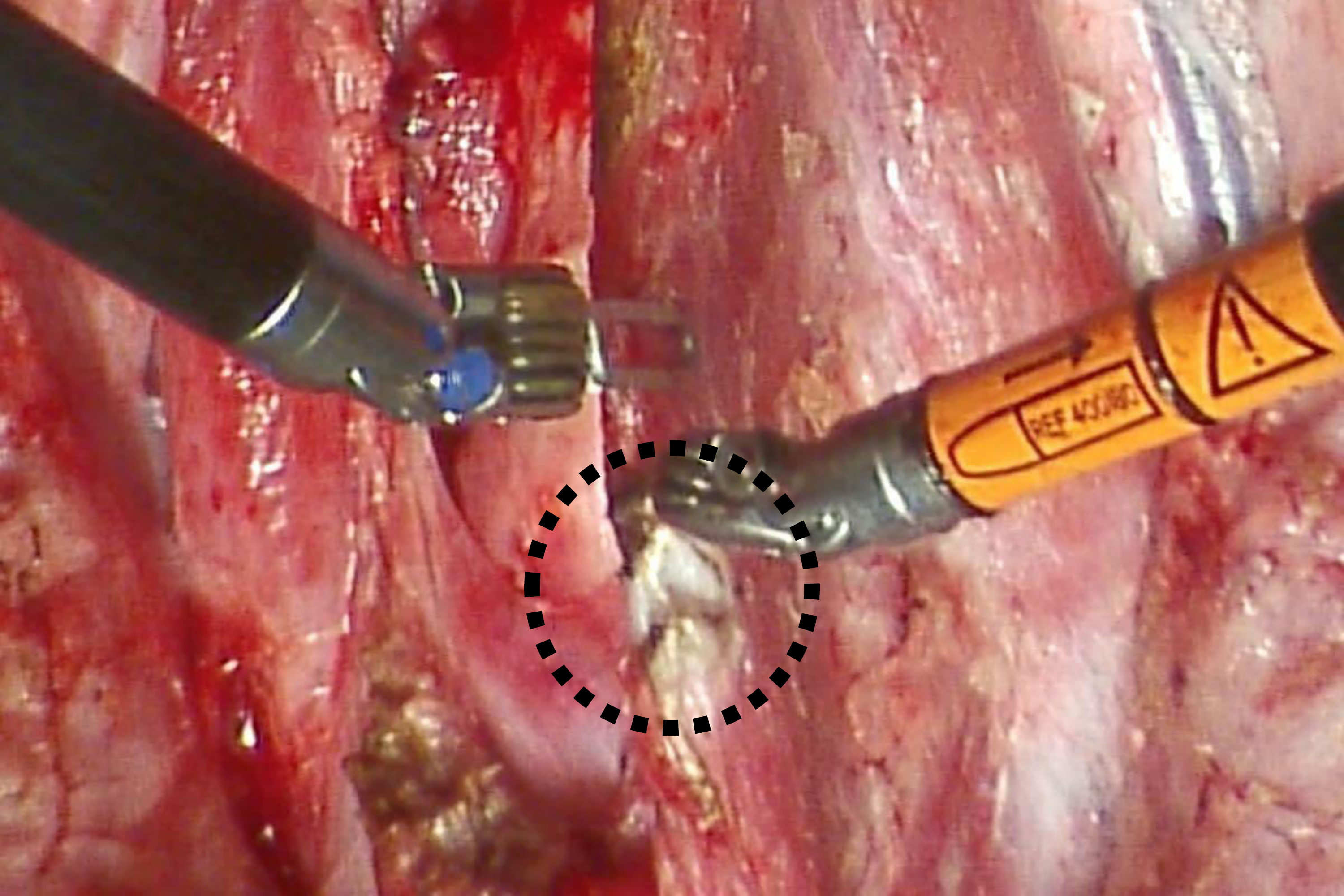}

    \caption{The figures from left to right are the first cut/cauterization during the sternohyoid resection under a round trip delay of 500msec with: normal constant scaling ($\gamma_c = 0.30, \gamma_v = 0.0$), reduced constant scaling ($\gamma_c = 0.15, \gamma_v = 0.0$), and velocity scaling ($\gamma_c = 0.15, \gamma_v = 0.1$). Notice the cuts with reduced constant scaling and velocity scaling are much deeper than normal constant scaling which results in faster dissection.}
    \label{fig:qualatative_results_cut}
\end{figure*}

During the live surgery under delay, the bedside assistant commented that there were a significant amount of uncontrolled motions when using the normal constant scaling under delay.
Meanwhile, when using the reduced constant scaling and velocity scaling methods, the motions looked substantially more controlled.
Qualitative examples of improved control when using reduced constant scaling and velocity scaling is shown in Fig. \ref{fig:qualatative_results_cut}.
The expert surgeon conducting the operation commented that operating under delay overall is very challenging and frustrating.    
Additionally, the reduced constant scaling was frustrating due how slow the motions on the bed-side-manipulators are.

\begin{figure}[t]
    \vspace{2mm}
    \centering
    \includegraphics[trim={11.5cm 10.2cm 0cm 0.65cm}, clip, width=0.9\linewidth]{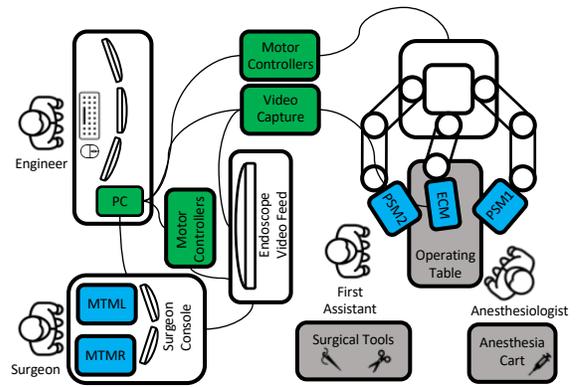}
    \caption{Schematic of the minimum configuration to conduct live surgeries for research. Highlighted in blue, grey, and green are the surgical robotic arms, electronics, and medical equipment respectively.}
    \label{fig:schematic}
\end{figure}


\section{Challenges with In-Vivo Experimentation}
A significant number of additional challenges and obstacles were overcome to get the presented in-vivo results.
The first of which is receiving approval to perform experimental procedures involving live animals.
The Institutional Animal Care and Use Committee (IACUC) at our institution serves to review all animal use protocols, ensuring compliance with federal regulations for the humane treatment and care of animals being used for testing, research and education.
The study presented here was approved under IACUC \# S19130.
While the protocol structure and application process varies between institutions, there are several important considerations when planning a live animal study.
First, there must be rationale and explanation for why live animals must be used in the study rather than alternative models.
The team must determine a reasonable number of required animals needed for the study, using the minimum possible.
The purpose and endpoints must be clear and provide insight on how the proposed research with ultimately benefit society, advance knowledge or benefit human or animal health.
The pre-operative care of the animals, anesthesia and medications to be used, surgical procedures, and post-operative care or method of euthanization must be clearly outlined.
Each surgical step must be broken down and described, including planned incisions, tissue removal and proposed length of surgery.
The researchers must provide detailed information on pain management and prevention, medication side effects, warming methods, disinfection and sanitation methods and intra-operative monitoring mechanisms to ensure animal safety.
All involved personnel must be listed and properly trained prior to protocol evaluation, including engineers who will be present for technical and robotic support.
While the IUCAC protocol is mainly the responsibility of the surgical members of the group, it is important to work closely with additional university faculty, including members of the Animal Care Program and the anesthesia team.

Another necessary component is finding and equipping an operating room for experimentation.
A schematic of the operating room is shown in Fig. \ref{fig:schematic}.
Surgical instruments and supplies include drapes, gauze, needles, syringes, protective equipment, suction supplies and sutures.
An electrocautery system was set up for use with the dVRK and handheld for hemostasis if needed.
A built-in suction port was set up and tested, and a V-shaped operating table was obtained for proper positioning of the animal in the supine position.
Monitors were installed to provide endoscopic video feed to the bedside assistant in the operating area.
On the day of the procedure, an anesthesia team provided the anesthesia machine, incubation supplies and all medications required for the induction and maintenance of anesthesia, as well as the agents used for euthanizing the animal after the procedure.

A major hurdle for the surgical team was adjusting to commanding the dVRK rather than the third or fourth generation Intuitive's da Vinci\textregistered{} Surgical Robot which are used in hospital operating rooms.
The joint limits and mobility of the MTMs were noted to be limited and forceful over-rotation or abrupt movement of the arms could cause the system to freeze.
The gravity compensation also seemed to be slightly less robust than the newer models.
Although seemingly minor, these differences required practice with the system in order to adapt to them.
In addition, the surgeons needed to be aware that bugs and delays may be encountered and require pauses in the procedure to resolve the technical issues.
Such events are not commonly encountered when using standard equipment in a hospital operating room and required the surgeon to alter their expectations to prevent frustration.
These adaptations and the previously described restrictions on ECM motion and surgical bedside assistance represent differences that are not standard in the operating room of a hospital.
However, they served to better evaluate our specific engineering goals and to maintain a replicable standard with as little variation as possible despite the relative complexity of the surgery.

\section{Discussion and Conclusion}
Although the results presented here provide a limited quantity of data points for validating motion scaling to enable remote telesurgery, it does show a proof of concept for live surgery implementations and the limited results do align with our previous laboratory experiments \cite{richter2019motion, orosco2020compensatory}. 
The motion scaling implementation on dVRK sends telecommands and visual feedback over network so the remote-administrator does not need to be local to the bed-side-manipulators.
The results also support our previous claim that velocity scaling produces more efficient motions through the reduced total time and distance travelled by the bed-side-manipulators to complete the surgical tasks.
Lastly, both reduced constant scaling and velocity scaling appear to have more controlled motions than normal constant scaling when under delay.

The live surgery experimentation described here is not just important for motion scaling and its progress to deployment in remote telesurgery, but also for the whole dVRK community.
This is the first live surgery conducted on a dVRK, hence being a big step forward to translating research from laboratory experiments to real surgeries for the entire community. 
By extension, it is the first live surgery conducted with an open-sourced robot, with hardware and software infrastructure co-developed by many individual contributors across the robotics research community.
The procedure and challenges with developing an in-vivo experiment are outlined here so others in the surgical robotics community can benefit from our experiences.
While the challenges to reach in-vivo experimentation are great, it is still a worthwhile endeavour since live studies create a true surgical environment which is less constrained than any other experimental model.
Furthermore, in-vivo surgical testing of an engineered solution brings a realm of new challenges that may highlight critical shortcomings, weaknesses, or even dangers of novel solutions.
Such challenges include movement relating to breathing, bleeding, different tissue properties, and robust muscle movement and rapid twitching with electrocautery or nerve stimulation.

For future work, we intend to use the live surgery experiment presented here with a larger number of randomized trials to further validate motion scaling for remote telesurgery.
Additionally, a scoring metric will be developed by expert surgeons to quantify performance of the surgical procedure from a clinical perspective.
This allows for a more robust method of evaluating performance of the surgical tasks from a clinical perspective.

\section{Acknowledgements}
This work is supported by University of California San Diego's ACTRI/IEM GEM, Health Science Bridge Award BG096652, Intuitive Surgical Technology Grant, the US Army TATRC under the Robotic Battlefield Medical Support System project, and NSF under grant number 1935329 and 2045803.
F. Richter is supported by the NSF Graduate Research Fellowship.
The authors would like to thank Intuitive Surgical Inc. for instrument donations, Simon DiMiao, Omid Maherari, Dale Bergman, and Anton Deguet for their support with the dVRK, and William Thomasson, Keith Jenne, and the animal care program at University of California San Diego for their assistance with the pigs and facility and equipment support.

\balance
\bibliographystyle{ieeetr}
\bibliography{references.bib}

\end{document}